\documentclass[10pt,twocolumn,letterpaper]{article}

\usepackage{cvpr}
\usepackage{times}
\usepackage{graphicx}
\usepackage{amsmath,amssymb} 
\usepackage{color}
\usepackage{subcaption}
\usepackage{booktabs}
\usepackage{algorithm}
\usepackage[noend]{algpseudocode}
\usepackage{multirow}
\usepackage{color, colortbl}
\usepackage{xspace, xcolor}

\definecolor{Gray}{gray}{0.9}

\newcommand{\overbar}[1]{\mkern 1.5mu\overline{\mkern-1.5mu#1\mkern-1.5mu}\mkern 1.5mu}

\def\algbackskip{\hskip-\ALG@thistlm}

\usepackage[pagebackref=true,breaklinks=true,letterpaper=true,colorlinks,bookmarks=false]{hyperref}

\cvprfinalcopy
\def\cvprPaperID{2554} 

\begin{document}

\title{Spatio-Temporal Instance Learning: Action Tubes from Class Supervision}

\author{Pascal Mettes and Cees G. M. Snoek\\\\University of Amsterdam}

\maketitle

\begin{abstract}
The goal of this work is spatio-temporal action localization in videos, using only the supervision from video-level class labels. The state-of-the-art casts this weakly-supervised action localization regime as a Multiple Instance Learning problem, where instances are \emph{a priori} computed spatio-temporal proposals. Rather than disconnecting the spatio-temporal learning from the training, we propose Spatio-Temporal Instance Learning, which enables action localization directly from box proposals in video frames. We outline the assumptions of our model and propose a max-margin objective and optimization with latent variables that enable spatio-temporal learning of actions from video labels. We also provide an efficient linking algorithm and two reranking strategies to facilitate and further improve the action localization. Experimental evaluation on four action datasets demonstrate the effectiveness of our approach for localization from weak supervision. Moreover, we show how to incorporate other supervision levels and mixtures, as a step towards determining optimal supervision strategies for action localization.
\end{abstract}

\section{Introduction}

This work focuses on the spatio-temporal localization of human actions such as \emph{Skiing} and \emph{Driving a car} when only video labels are given during training. Since box annotations are cumbersome, tedious, and error-prone to annotate~\cite{manen2017pathtrack,mettes2018pointly}, weakly-supervised works aim to localize actions from class supervision~\cite{chen2015action,li2018videolstm,sapienza2012learning,sharir2014action}, optionally supplemented with actor detectors~\cite{mettes2017localizing,siva2011weakly}. The state-of-the-art in weakly-supervised action localization~\cite{mettes2018pointly,mettes2017localizing,sapienza2012learning} splits videos \emph{a priori} into spatio-temporal proposals~\cite{jain2017tubelets,oneata2014spatio,vangemert2015apt} and casts the localization as a Multiple Instance Learning problem.
Hence, learning amounts to finding and using the best proposal per training video.
This approach has two limitations: (1) the spatio-temporal localization is disconnected from the learning and (2) spatio-temporal proposals are not competitive for action localization~\cite{mettes2018pointly}. In this work, we propose to bypass the spatio-temporal proposals and learn the spatial and temporal extent of actions during training.

\begin{figure}[t]
\centering
\includegraphics[width=\linewidth]{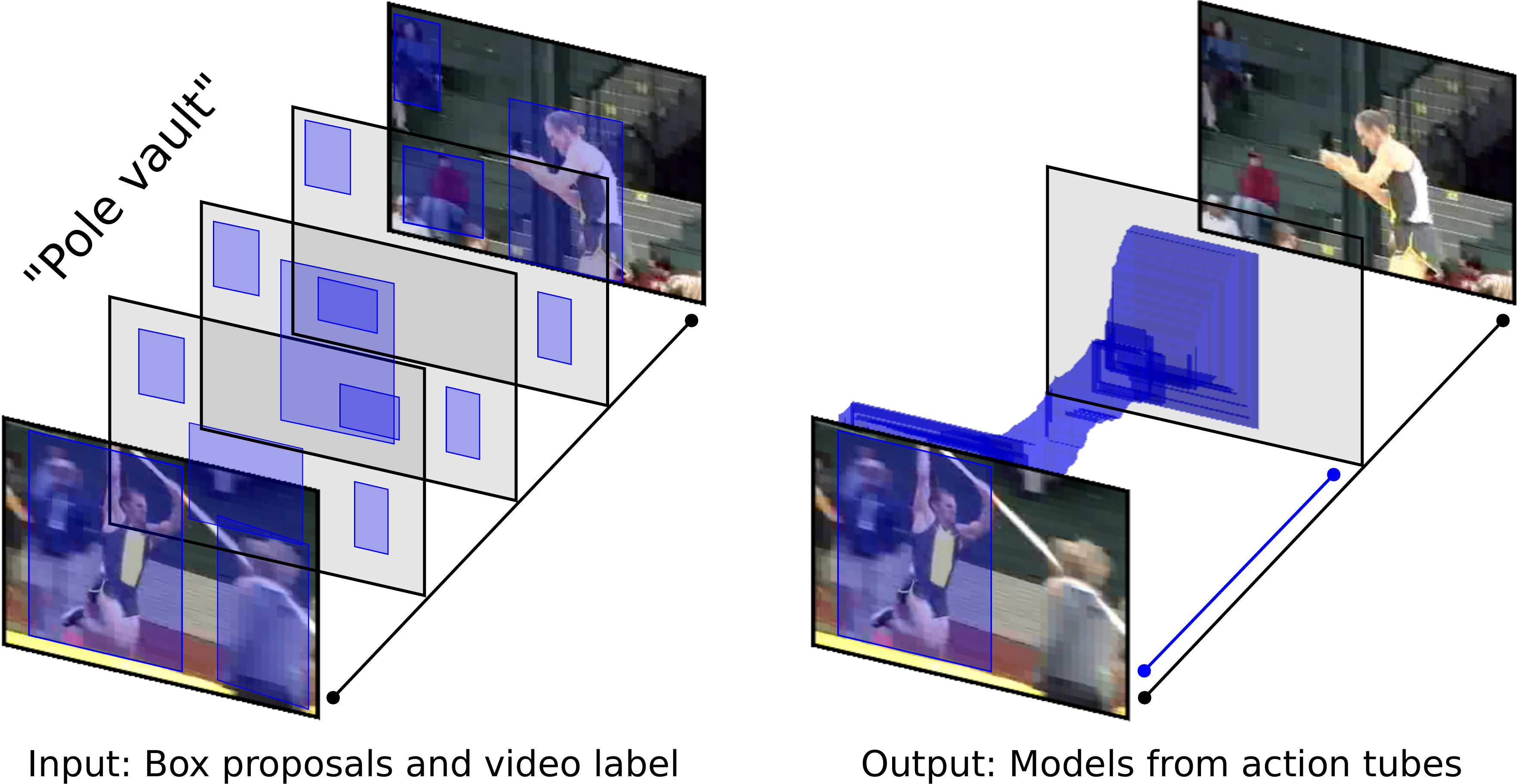}
\caption{\textbf{Illustration of Spatio-Temporal Instance Learning.} Given only video labels and box proposals in frames, our approach discovers spatio-temporal action tubes that enable the training of action localization models.}
\label{fig:fig1}
\end{figure}

We are inspired by the recent success in action localization \emph{with} ground truth box supervision. While these works initially also profited from spatio-temporal proposals prior to learning~\cite{jain2017tubelets,oneata2014spatio,vangemert2015apt}, Gkioxari and Malik~\cite{gkioxari2015finding} showed that spatial action detection per frame before temporal linking is beneficial for action localization. This line of work resulted in further progress with better deep representations~\cite{peng2016multi,yang2017spatio}, better temporal linking~\cite{singh2017online}, or both~\cite{weinzaepfel2015learning,saha2016deep}.
Most recent works either classify up to ten consecutive frames before linking~\cite{hou2017tube,kalogeiton2017action,saha2017amtnet,singh2018tramnet} or add recurrency after linking boxes into tubes~\cite{he2018generic,li2018recurrent} to obtain state-of-the-art localization from box supervision. We adopt the principle of separating the spatial action classification from the temporal linking, but rather than relying on ground truth box supervision, we do so from a weakly-supervised signal.

We propose Spatio-Temporal Instance Learning (STIL), an algorithm for action localization from video labels, see Figure~\ref{fig:fig1}. Rather than providing spatio-temporal instances \emph{a priori}, as with Multiple Instance Learning, we aim to learn a model from box proposals in frames, while adhering to the spatio-temporal nature of actions. We first define three model assumptions and introduce a latent instance learning function to train box-level models with spatio-temporal consistency. To facilitate the spatio-temporal learning, we propose an efficient procedure for linking box proposals into tubes from zero-shot priors, \ie without the need for ground truth boxes. From trained models, we score and link box proposals in test videos for an effective action localization. To further improve the localization, we provide a reranking of test tubes based on contextual information and information from other localized actions. Lastly, we show how to incorporate action supervision beyond class labels to enhance the localization. Experimental evaluation on four action localization datasets shows the effectiveness of our proposal for weakly-supervised action localization.


\section{Related work}
Multiple Instance Learning (MIL) is an algorithm for supervised learning, where annotations are not provided for each instance, but only for bags of instances~\cite{dietterich1997solving,maron1998framework}. The main assumptions in this regime are that a positive bag should contain at least one positive instance, while a negative bag should contain only negative instances. Generalized variants of MIL~\cite{li2015multiple,scott2005generalized} presume the presence of more than one positive instance per bag, for problems such as content-based image retrieval. Both MIL and its generalized variants typically focus on classification at the bag level~\cite{andrews2003support,dietterich1997solving,maron1998framework,vanwinckelen2016instance}. In this work, we introduce a novel instance learning model, dubbed Spatio-Temporal Instance Learning, that aims for the spatio-temporal learning of actions in videos, rather than classifying whole videos.

For localization in videos and images from global labels, MIL is commonly used. To fit the localization problem to MIL, instances are generated \emph{a priori} in the form of proposals. Cinbis \etal~\cite{cinbis2017weakly} perform object localization in images with MIL using object proposals~\cite{uijlings2013selective}. For the problem of spatio-temporal action localization, Siva and Xiang~\cite{siva2011weakly} propose a similar MIL approach in the video domain by splitting videos into action proposals derived from person detection, while Sapienza \etal~\cite{sapienza2012learning} rely on spatio-temporal cuboids. Mettes \etal~\cite{mettes2017localizing} also employ MIL on action proposals generated by clustering dense trajectories~\cite{vangemert2015apt}. Chen and Corso~\cite{chen2015action} skip the MIL step by directly selecting their most dominant action proposal during training. Yan \etal~\cite{yan2017weakly} use \emph{a priori} computed action proposals with semantic label inference to segment both actions and actors using video labels. Li \etal~\cite{li2018videolstm} perform action localization by learning and linking visual attention in video frames~\cite{sharma2015action}. They localize actions in frames with a box around the center of attention and perform linking \emph{a posteriori} into action tubes. Rather than separating the spatio-temporal localization from the learning, we propose a MIL variant that integrates the localization during the learning from video labels.

Beyond video labels, a few works have included other forms of spatio-temporal supervision for action localization. Mettes \etal~\cite{mettes2018pointly} start from MIL with spatio-temporal proposals and add manual point supervision to guide the selection of proposals during training. Weinzaepfel \etal~\cite{weinzaepfel2017human} start from person tube proposals and select based on one to five manual box annotations together with the temporal action span. A challenge for such supervision tactics is the difficulty to compare them on equal grounds, \ie using same visual features and learning scheme.
We show how to incorporate additional levels of supervision in our model and use this to directly compare their localization performance as a function of their annotation time.

A number of works have recently investigated action localization without any video examples, where objects are typically employed to localize actions~\cite{kalogeiton2017joint,mettes2017spatial}. Soomro and Shah~\cite{soomro2017unsupervised} aim for action localization in videos without any annotations through discriminative clustering and by linking supervoxels with 0-1 Knapsack. While promising, these approaches are currently not competitive to supervised alternatives in action localization and we will therefore not compare to them in this work.

Related to our setting is the problem of weakly-supervised temporal action detection. Such works include learning the temporal extent of actions from video labels~\cite{nguyen2018weakly,paul2018w,shou2018autoloc,wang2017untrimmednets} or learning the change of actions from the order of action labels in videos~\cite{bojanowski2014weakly,huang2016connectionist}. Akin to these works, we aim to discover the extent of actions from weak supervision (video labels only), but we focus on the extent in space and time, rather than time only.

\section{Spatio-Temporal Instance Learning}
In our weakly-supervised setting, we are given $N$ training videos, where each video is represented by a set of automatically detected box proposals in individual frames and a global video label. We use the box proposals to decompose training video $i$ as $\mathbf{x}_i \in \mathbb{R}^{T \times F_i \times D}$, where $T$ denotes the number of tubes, $F_i$ denotes the number of frames, and $D$ denotes the feature dimensionality of a box.
Before detailing how to obtain such a decomposition when only global video class supervision is given, we first detail the conditions, objective function, and optimization of Spatio-Temporal Instance Learning itself.
\subsection{Model}
\label{sec:model}

\noindent
\textbf{Conditions.}
For an action class, training videos are split into positive and negative videos. We perform action localization using just the video labels as supervision. Akin to standard Multiple Instance Learning (MIL)~\cite{dietterich1997solving,maron1998framework}, we impose several conditions on the videos. The first condition is given as:
\\\\
\textbf{Condition 1:}
\emph{Each positive video contains at least one positive action instance, which can occur in precisely one tube.}
\\\\
The first part of the condition follows MIL, in that we can only guarantee the presence of one instance in each video, as further supervision is not provided. The second part considers the spatial extent of actions; an action can only occur in one location and this location can be tracked over time. The second condition is given as:
\\\\
\textbf{Condition 2:}
\emph{For each positive video $V$, the positive action instance is a set of connected boxes of minimal length $1$ and maximal length $F_V$, where $F_V$ denotes the total video length.}
\\\\
This condition considers the temporal extent of actions; an action can be of any length, but its occurrence is continuous over time. Lastly:
\\\\
\textbf{Condition 3:}
\emph{For each negative video, all tubes and boxes are negative.}
\\\\
The third condition provides the constraints for negative videos, for which the spatio-temporal extent should not resemble positive videos at all.
While the conditions mostly follow common intuitions about the spatio-temporal extent of actions in videos, they are not covered in standard MIL. Moreover, they provide the constraints to shape our objective function and optimization.
\\\\
\noindent
\textbf{Objective function.}
Given an action $c$, let us denote the training set as $\{\mathbf{x}_i, y_i^c\}_{i=1}^{N}$, where $y_i^c \in \{-1,+1\}$ denotes the binary video label with respect to $c$. We aim for a classifier $(\mathbf{w},b)$ to localize actions at the box-level based on the provided STIL conditions. We propose an instance learning objective function able to maximize localization performance, guided by the conditions. The full objective function is given as:
\begin{equation}
\begin{split}
\label{eq:objective1}
\min_{\mathbf{w},b,\mathbf{h}} \frac{1}{2} ||\mathbf{w}||^{2} + & C \sum_{i=1}^{N} \sum_{j=1}^{T} \sum_{k=1}^{F_i} h_{ijk} \cdot \\
 & \text{max} \big[0, 1 - y_i^c \cdot (<\!\! \mathbf{w}, \mathbf{x}_{ijk} \!\!> + b) \big],\\
\end{split}
\end{equation}
\begin{equation}
\begin{split}
\label{eq:objective2}
\text{s.t.} \quad & \forall_{ijk} : h_{ijk} \in \{0,1\},\\
& \forall_{i} : 1 \leq \sum_{j,k} h_{ijk} \leq F_i,\\
& \forall_{i} : \sum_{j} [\![ \sum_{k} h_{ijk} > 0 ]\!] = 1,\\
& \forall_{ij} : \sum_{k} [\![ h_{ijk} \neq k_{ijk+1} ]\!] \leq 2,
\end{split}
\end{equation}
where $[\![ \cdot ]\!]$ denotes the Iverson bracket and $C$ denotes the regularization parameter.
Eq.~\ref{eq:objective1} states a max-margin objective, there the hinge loss is computed over all boxes of all tubes of all training videos. Each box representation $\mathbf{x}_{ijk}$ is accompanied by a binary latent variable $h_{ijk}$, which determines whether the box should contribute to the loss. The constraints on the binary latent variables are given in Eq.~\ref{eq:objective2}. The first constraint states the binary nature of the latent variables. The second constraint states the temporal extent of actions in each tube (condition 2). The third constraint states that at only one tube in each video has boxes that contribute to the loss (condition 1). The fourth constraint states that the boxes that contribute to the loss are consecutive within a tube (condition 2). Note that condition 3 is automatically incorporated since the top scoring boxes of negative videos are treated as negative examples in the hinge loss. As such, our proposed objective function incorporates the conditions through a max-margin formulation with constrained binary latent variables.
\\\\
\noindent\textbf{Optimization.}
To solve our objective function, we rely on Expectation-Maximization. This optimization alternates between determining the values of the latent variables (expectation) and solving the max-margin objective (maximization). Given fixed latent variables, the maximization step can be tackled with standard $\ell_2$-regularized max-margin solvers~\cite{fan2008liblinear}.
The expectation step revolves around assigning binary values to the latent variables subject to the constraints from Eq.~\ref{eq:objective2}. To that end, we define a score function for each subtube in a video $i$ as:
\begin{equation}
z(\mathbf{x}_i, j, k, s) = \bigg( \sum_{l=k}^{k+s} (<\!\! \mathbf{w}, \mathbf{x}_{ijl} \!\!> + b) \bigg) - s,
\label{eq:estep}
\end{equation}
where $j, k, s$ denote the tube index, starting box index, and stride respectively. The first part of Eq.~\ref{eq:estep} states the sum of scores over the boxes in the subtube with the current model parameters. The second part states a regularization on the subtube length. We find the subtube with the highest regularized score and set the latent variables that correspond to that subtube to one. The length regularization aims to balance the precision versus recall. Given a subtube per training video, we use the updated latent variables to retrain the model. This process is iterated multiple times to refine the model further. The process is initialized by replacing the model scores in Eq.~\ref{eq:estep} with the prior box scores.

\subsection{Linking boxes with Temporal Prim}
\label{sec:linking}
To uncover the spatio-temporal extent of actions, we enforce a decomposition of videos into video-length tubes, which in turn consist of boxes. Since full box supervision is not provided, the tubes are not \emph{a priori} known in our setting. Here, we propose an efficient algorithm to automatically obtain tubes in videos. For a box $B = (B^{f}, B^{s})$ with frame and spatial coordinates, we first obtain a zero-shot prior score $s(B)$, \eg using spatial-aware embeddings from~\cite{mettes2017spatial}. In practice, we found that using an actor detector is sufficient and we use this throughout our work. With a score for each box proposal in a video, we obtain a small set of action tubes that densely cover the actions in a video.
To obtain action tubes, we introduce a temporal variant of Prim's algorithm~\cite{prim1957shortest}, dubbed Temporal Prim, where we start from any actor box and grow tubes efficiently through the video. The box proposals are first connected into a sparsely connected graph. For boxes $B_i$ and $B_j$, the edge weight is determined as:
\begin{equation}
e(B_i, B_j) =
\begin{cases}
s(B_i) + s(B_j) &  \text{if } |B_i^f - B_j^f| = 1 \text{ and}\\
& \text{iou}(B_i^s, B_j^s) \geq 0.1,\\
0 & \text{otherwise.}
\end{cases}
\label{eq:primedge}
\end{equation}
where $\text{iou}$ denotes the spatial overlap function. Eq.~\ref{eq:primedge} states that non-zero edges only exist between boxes of consecutive frames with an overlap of at least 0.1. We start Temporal Prim from the box with the highest overall zero-shot prior score. We then evaluate two sets of edges, one forward in time, one backward. We extend our tube backward or forward in time, depending on which edge has the highest weight overall. The box connected to the edge is added to the
tube. From the newly selected box, we look one frame further ahead or back (depending on the direction in which we moved), and compute new edge weights between the boxes from that frame and the selected box. We remove the old edge weight connected to the selected box and replace them with the new edge weights. We continue the procedure of adding boxes and updating edge weights until the start/end of the video is reached, or no boxes with prior support could be found.
\\\\
\textbf{Complexity analysis.} Temporal Prim exploits the knowledge that an action only moves along the temporal axis (\ie forward and backwards). This allows us to reduce the computational complexity in the Prim algorithm. Let $\overbar{b}$ denote the (upper bound of the) number of boxes per frame and let $F_i$ denote the number of frames in video $i$. Our approach has a computational complexity of $\mathcal{O}(1 + \overbar{b})$ to add a new box to a tube ($\mathcal{O}(1)$ for determining direction, $\mathcal{O}(\overbar{b})$ to overwrite and instantiate the set of edges in the determined direction). Generating a video-length tube has a computational complexity of $\mathcal{O}(F_i \cdot \overbar{b})$. By exploiting the bidirectional nature of action tubes, our Temporal Prim has a lower computational complexity than its spatial~\cite{manen2013prime} and spatio-temporal~\cite{oneata2014spatio} (randomized) variants, which have a complexity of $\mathcal{O}((F_i \cdot \overbar{b}) \cdot \text{log}(F_i \cdot \overbar{b}))$ ~\cite{manen2013prime}.
\\\\
\textbf{Relation to other temporal linkers.}
Several works in action localization \emph{with} box supervision have proposed linkers to connect boxes into tubes.
These linkers typically focus on generating tubes at test time from actions scores, as tubes are readily provided during training. The linkers of Gkioxari and Malik~\cite{gkioxari2015finding} and Weinzaepfel \etal~\cite{weinzaepfel2015learning} for example start at the first frame and therefore require a box annotation for initialization. The recent linkers of Singh \etal~\cite{singh2017online} and Behl \etal~\cite{behl2018incremental} aim for online linking of boxes of all actions simultaneously using models trained on box annotations. Since we aim for action localization from video labels, a starting box annotation is absent and the highest action prior can be anywhere in the video. Furthermore, since we require action tubes in training videos, simultaneous linking of all actions from learned models is not yet possible.

\subsection{Beyond class label supervision}
\label{sec:beyond}
The aim of Spatio-Temporal Instance Learning is to uncover the spatio-temporal extent of actions when only global video labels are provided. Given the general nature of our objective and linking algorithm, we are able to incorporate additional supervision signals, beyond video labels, to guide the learning process. We focus here on three supervision signals in frames: box annotations, point annotations, and no annotation. For a box $B$, we define the following scoring function for additional supervision:
\begin{equation}
s(B, \mathcal{A}_i) = \sum_{j} \delta_t(\mathcal{A}_{ij}^{f}, B^{f}) \cdot \delta_s(\mathcal{A}_{ij}^{s}, B^{s}),
\label{eq:extrasup}
\end{equation}
where $\delta_t$ and $\delta_d$ denote the temporal and spatial score function for annotations $\mathcal{A}_i$ in video $i$. The temporal score function is identical for all supervision signals:
\begin{equation}
 \delta_t(\mathcal{A}_{ij}^{f}, B^{f}) = \max \bigg(0, 1 - \frac{|\mathcal{A}_{ij}^{f} - B^{f}|}{2} \bigg).
\end{equation}
The temporal score function states that an annotation should not only matter for the frame in which it is annotated, but also for the two neighbouring frames, albeit with half the score. The spatial score function is given as:
\begin{equation}
\delta_s(\mathcal{A}_{ij}^{s}, B^{s}) =
  \begin{cases}
    \text{iou}(\mathcal{A}_{ij}^{s}, B^{s}) & \text{if } \mathcal{A}_{ij}^t \text{ is box,}\\
    \text{pmatch}(\mathcal{A}_{ij}^{s}, B^{s}) & \text{if } \mathcal{A}_{ij}^t  \text{ is point,}\\
    -1 & \text{if } \mathcal{A}_{ij}^t \text{ is none,}
  \end{cases}
\end{equation}
with $\text{pmatch}(\mathcal{A}_{ij}^{s}, B^{s}) = \max(0, 1 - \frac{||\mathcal{A}_{ij}^s - c(B^s)||}{||c(B^s) - e(B^s)||})$. The spatial score function depends on the annotation type. For box annotations, we use the spatial overlap function. For point annotations, we use the overlap function given by~\cite{mettes2018pointly}. Lastly, we add a negative score for boxes in frames where the action is absent. In STIL, we add the scores from Eq.~\ref{eq:extrasup} from additional annotations to the prior scores of the box proposals in training videos. This in turn leads to a max-margin Expectation-Maximization with better initialization, resulting in more precise action localizers.

\subsection{Inference}
\label{sec:inference}
At test time, we first score each box proposal in each test video using our trained action localization model for each action separately. Then, we apply Temporal Prim on the boxes, now using the model scores instead of the prior scores. We stop extending tubes if no support is found in new frames (\ie if no boxes with positive action scores are present), arriving at spatio-temporal tubes in test videos. Let $(t_j^{\text{start}},t_j^{\text{end}})$ denote start and end frames for tube $j$. Then for action parameters $(\mathbf{w}_c, b_c)$, we score each tube $j$ in video $i$ as: $f(\mathbf{x}_i, j, \mathbf{w}_c, b_c) = \frac{1}{|t_j^{\text{end}}-t_j^{\text{start}}|} \sum_{k=t_j^{\text{start}}}^{t_j^{\text{end}}} (<\!\! \mathbf{w}_c, \mathbf{x}_{ijk} \!\!> + b_c)$. We use the scores to rank the spatio-temporal tubes found in all test videos and arrive at an action localization from box-level features. To further improve this localization, we investigate two reranking strategies based on information from context and from other actions.
\\\\
\noindent
\textbf{Contextual reranking.}
While performing action localization from box-level information has shown to be effective~\cite{gkioxari2015finding,weinzaepfel2015learning}, it ignores contextual information from the rest of the video that can help to discriminate tubes from different videos. Here, we incorporate this information by learning a global video classifier from the same training videos. For each tube, we simply add the score from the whole video given the global classifier to the tube score to rerank the tubes across all test videos.
\\\\
\textbf{Negative evidence reranking.}
We also exploit knowledge about other actions to rerank tubes in test videos. For a test video $i$ and action $c$, it is intuitive that the presence of other high-scoring actions has a negative impact on the presence of action $c$. Or conversely, the lower the score of all other actions in video $i$, the higher the rank of tubes for action $c$. Here, we incorporate the notion of negative evidence reranking by subtracting the difference between the maximum tube score over all actions and the tube score. The higher the difference, the larger the penalty.

\section{Experimental setup}

\noindent
\textbf{Datasets.}
We evaluate our approach on the four datasets common in the weakly-supervised action localization literature~\cite{chen2015action,li2018videolstm,mettes2017localizing,sharma2015action}. \emph{UCF Sports} consists of 150 sports videos from 10 actions~\cite{rodriguez2008action}. We employ the train/test split provided by~\cite{lan2011discriminative}. \emph{J-HMDB} consists of 928 videos from 21 daily activities, where we use the train/test split provided by~\cite{jhuang2013towards}. \emph{UCF-101-24} is a subset of UCF-101 with localization annotation. The dataset contains 24 actions from 3,207 videos, where we use the first split as provided by~\cite{soomro2012ucf101}. Lastly, \emph{Hollywood2Tubes} contains action localization annotations~\cite{mettes2016spot} for the videos from Hollywood2~\cite{marszalek09}. The dataset contains 1,707 videos from 21 actions and we use the train/test split provided by~\cite{marszalek09}.
\\\\
\textbf{Implementation details.}
We use Faster R-CNN~\cite{ren2015faster} pre-trained on person images~\cite{weinzaepfel2017human} to obtain the box proposals and prior scores in video frames. The ROI-pooling layer of the network yields a 4,096D feature representation per box. We $\ell_2$-normalize each box representation. For UCF-101 and Hollywood2Tubes, we perform box extraction every $5^{th}$ frame, while we sample every frame for UCF Sports and J-HMDB.
In STIL, we fix the regularization parameter $C$ to 10 in our experiments and train for 5 epochs. For the video features, we sample 2 frames per second and feed the frames to a pre-trained CNN~\cite{mettes2016imagenet}, followed by average pooling and $\ell_2$-normalization. Following~\cite{cinbis2017weakly}, we split our instance learning into 3 folds and we add background boxes from each video as additional negatives.
\\\\
\textbf{Evaluation protocol.}
For an action tube $a$ and ground truth $B$ from the same video, their spatio-temporal (st) overlap is given as: $\text{st-iou}(a,b) = \frac{1}{|\Gamma|} \text{iou}_f(a,b)$, where $\Gamma$ denotes the union of frames in $a$ and $b$. For localization with overlap $\tau$, an action tube is positive if the tube is from a positive video, the overlap with a ground truth instances is at least $\tau$, and the ground truth instance has not been detected before. We use mAP and AUC for evaluation.

\section{Experimental results}

\subsection{Ablation studies and analysis}

\noindent

\begin{figure*}[t]
\centering
\includegraphics[width=0.22\linewidth]{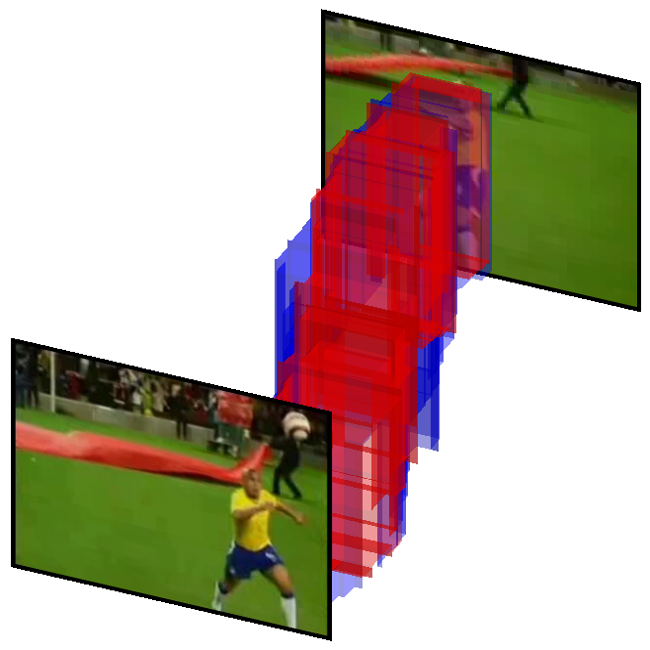}
\includegraphics[width=0.22\linewidth]{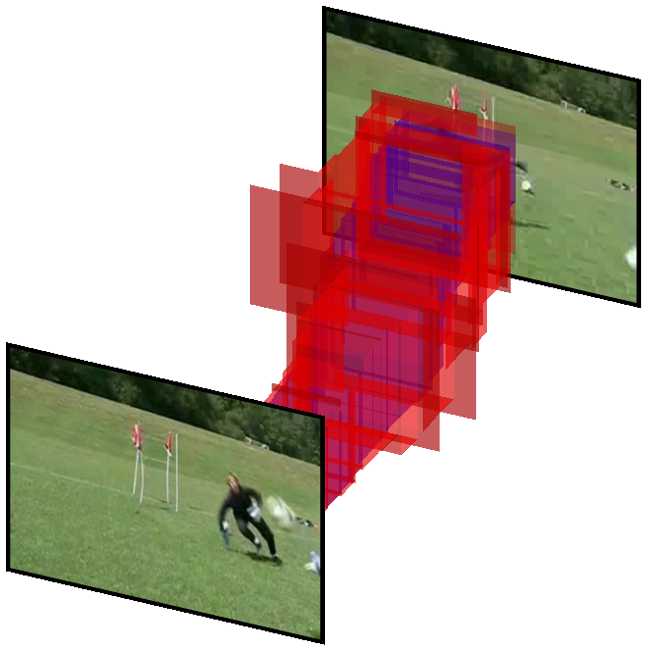}
\includegraphics[width=0.22\linewidth]{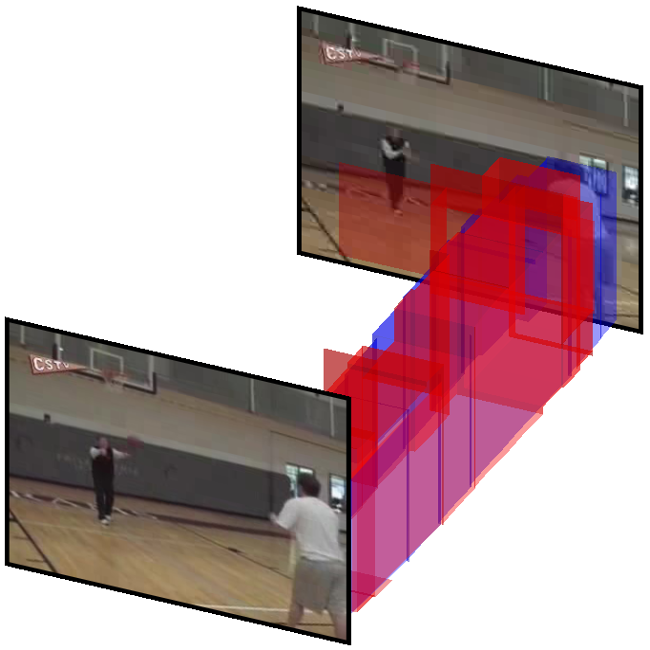}
\includegraphics[width=0.22\linewidth]{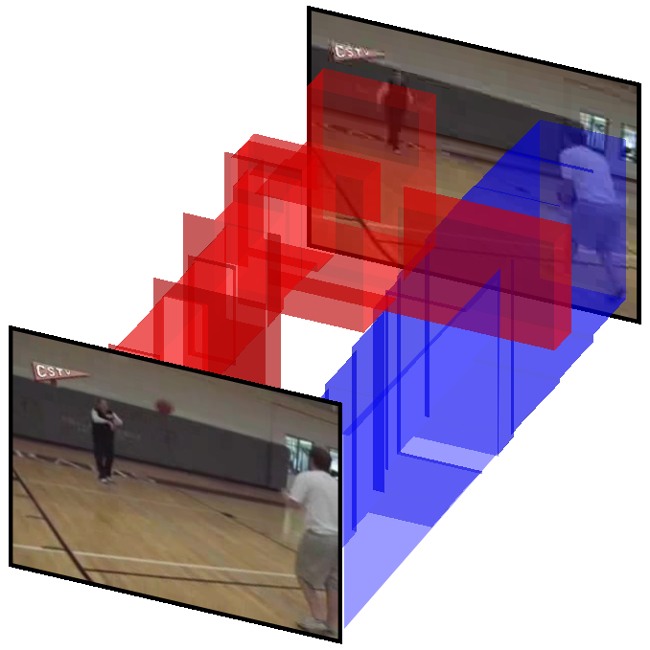}
\caption{\textbf{Qualitative results of STIL} on four test videos of J-HMDB, for the actions \emph{Kick}, \emph{Jump}, and \emph{Catch} (twice). Blue denotes the ground truth, red our predicted tube. The first two results have a high overlap with the ground truth. For \emph{Catch} with two test videos of the same scene, our approach captures the action in the first example, but confuses the action in the second example with the action \emph{Throw}, highlighting the difficulty of weakly-supervised action localization for videos with multiple concurrent actions.}
\label{fig:qual}
\end{figure*}
\begin{table}[t]
\centering
\resizebox{\columnwidth}{!}{%
\begin{tabular}{lccc}
\toprule
 & \multicolumn{3}{c}{\textbf{mAP@0.5}}\\
 & \footnotesize{UCF Sports} & \footnotesize{J-HMDB} & \footnotesize{UCF-101-24}\\
\midrule
\rowcolor{Gray}
\textbf{Baselines} & & &\\
MIL & 0.20 & 0.08 & 0.01\\
Generalized MIL ($\mu$=5) & 0.17 & 0.11 & 0.02\\
Generalized MIL ($\mu$=10) & 0.29 & 0.11 & 0.02\\
Generalized MIL ($\mu$=40) & 0.28 & 0.10 & 0.02\\
\midrule
\rowcolor{Gray}
\textbf{This paper} & & &\\
STIL (w/o Temporal Prim) & 0.63 & 0.27 & 0.05\\
STIL (w Temporal Prim) & \textbf{0.72} & \textbf{0.30} & \textbf{0.08}\\
\bottomrule
\end{tabular}%
}
\caption{\textbf{The impact of STIL and Temporal Prim} on the action localization performance across UCF Sports, J-HMDB, and UCF-101-24. For the Generalized MIL baselines, $\mu$ denotes the number of boxes used per video for training. Our approach outperforms the (Generalized) MIL and also STIL without temporal linking, highlighting the effectiveness of our spatio-temporal learning.}
\label{tab:exp1-1}
\end{table}
\noindent
\textbf{The impact of STIL.}
We first evaluate the effect of STIL with respect to standard instance learning. We compare to four baselines. The first baseline performs Multiple Instance Learning (MIL) on the boxes in each training video, \ie using a single positive box per video for training. We also compare to three Generalized MIL baselines, where a fixed number of boxes $\mu$ is used per video ($\mu$ is 5, 10, and 40 in the experiments). For initialization of the baselines, we perform random box selection. For all baselines and our approach, we use the same max-margin optimization, the approaches only differ in which boxes are used for training.
\begin{figure}[t]
\centering
\includegraphics[width=\linewidth]{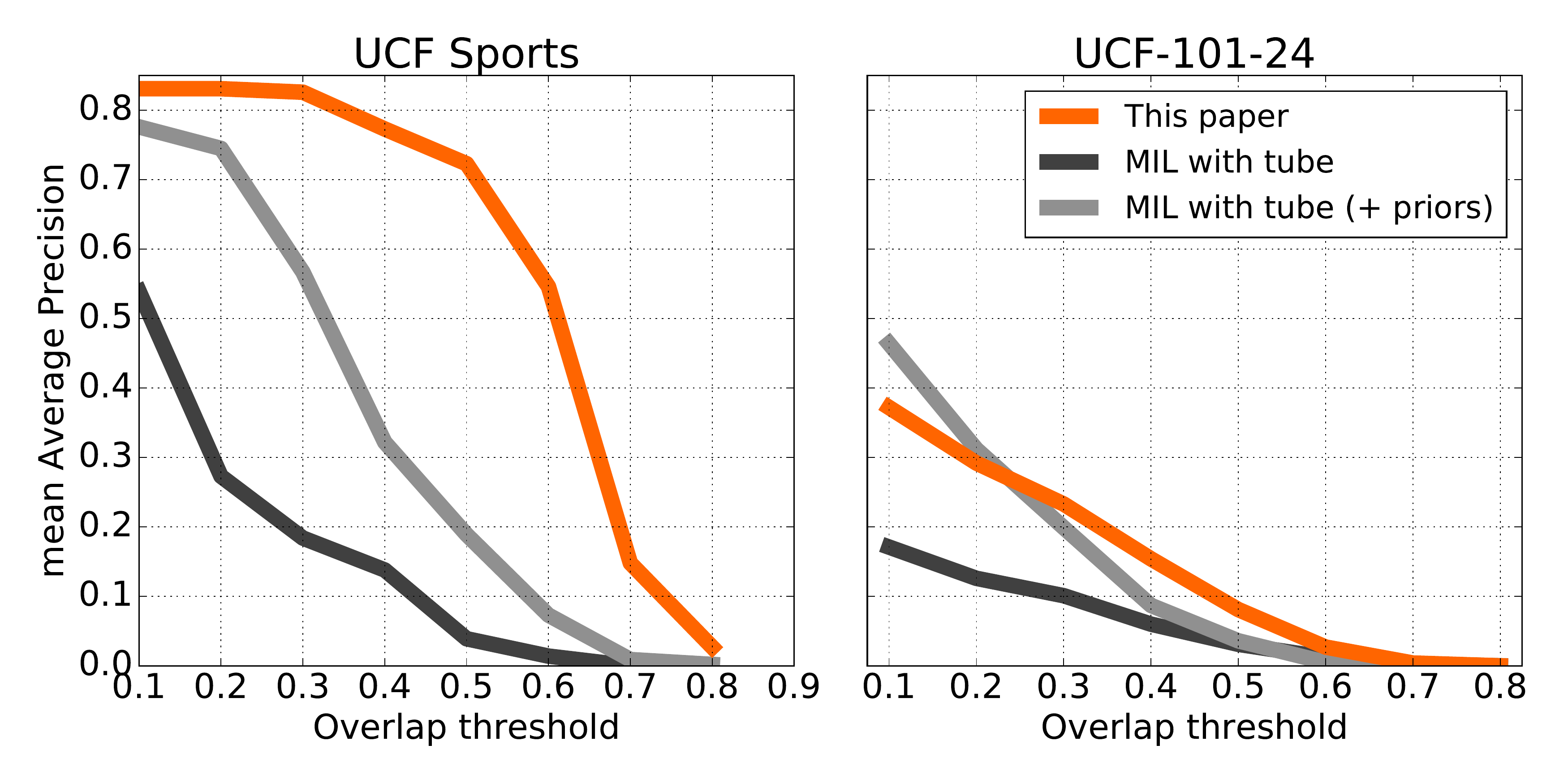}
\caption{\textbf{STIL on boxes versus MIL on tubes} for UCF Sports and UCF-101-24. Our approach outperforms the MIL baselines using spatio-temporal tube proposals, especially at higher overlap thresholds. STIL on boxes is preferred over MIL with tubes.}
\label{fig:exp1-3}
\end{figure}
\begin{figure*}[t]
\centering
\begin{subfigure}{0.3\linewidth}
\includegraphics[width=\linewidth]{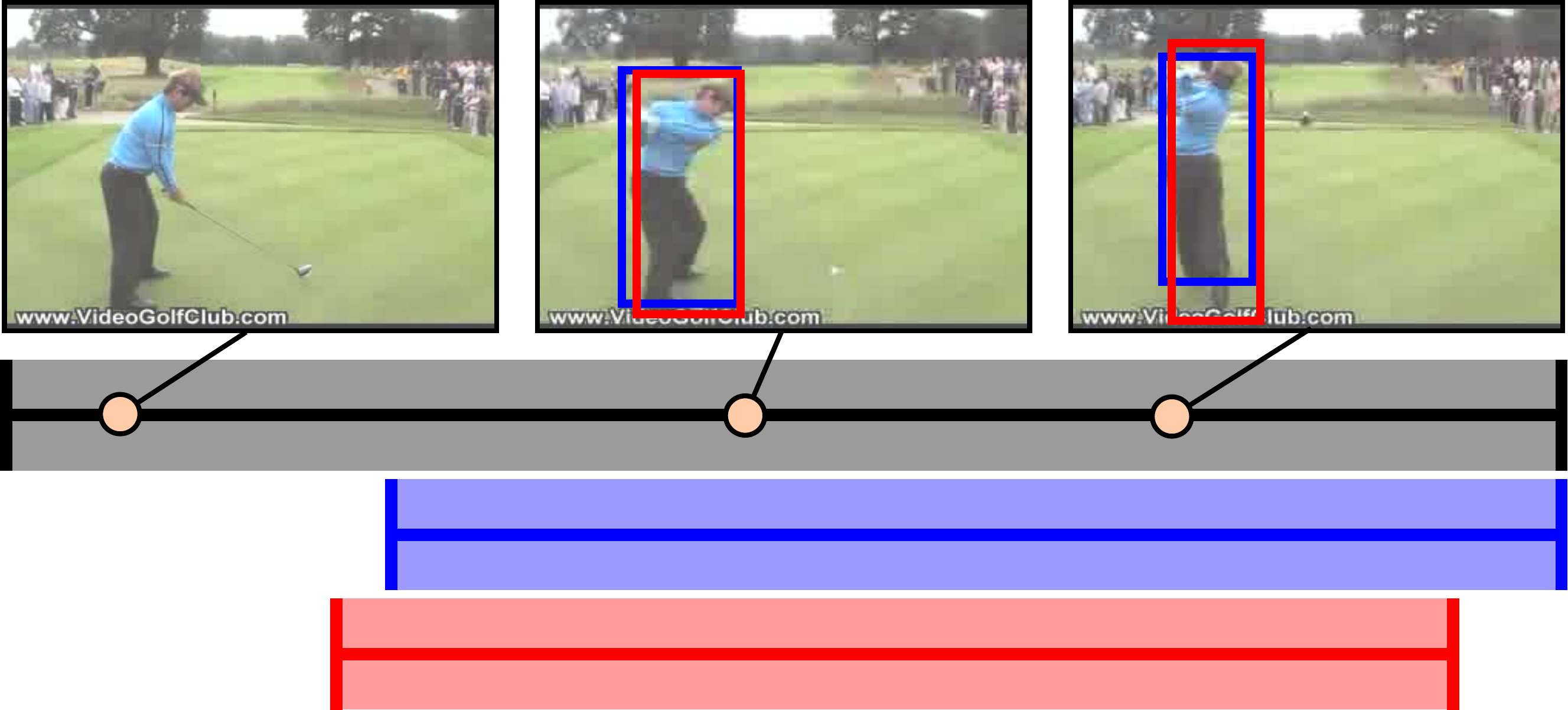}
\caption{\emph{Golf swing.}}
\label{fig:qual-ucf10124-a}
\end{subfigure}
\hspace{0.25cm}
\begin{subfigure}{0.3\linewidth}
\includegraphics[width=\linewidth]{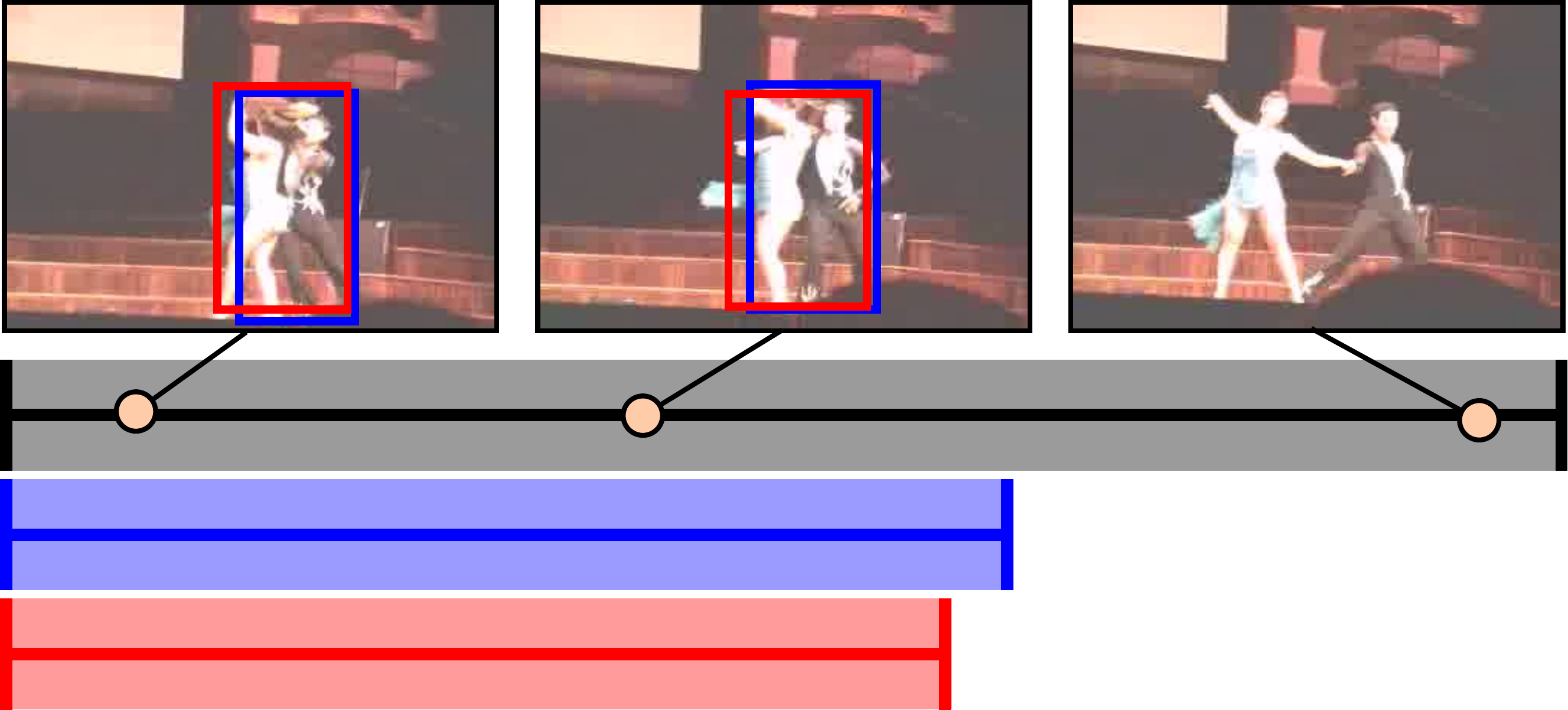}
\caption{\emph{Salsa spin.}}
\label{fig:qual-ucf10124-b}
\end{subfigure}
\hspace{0.25cm}
\begin{subfigure}{0.3\linewidth}
\includegraphics[width=\linewidth]{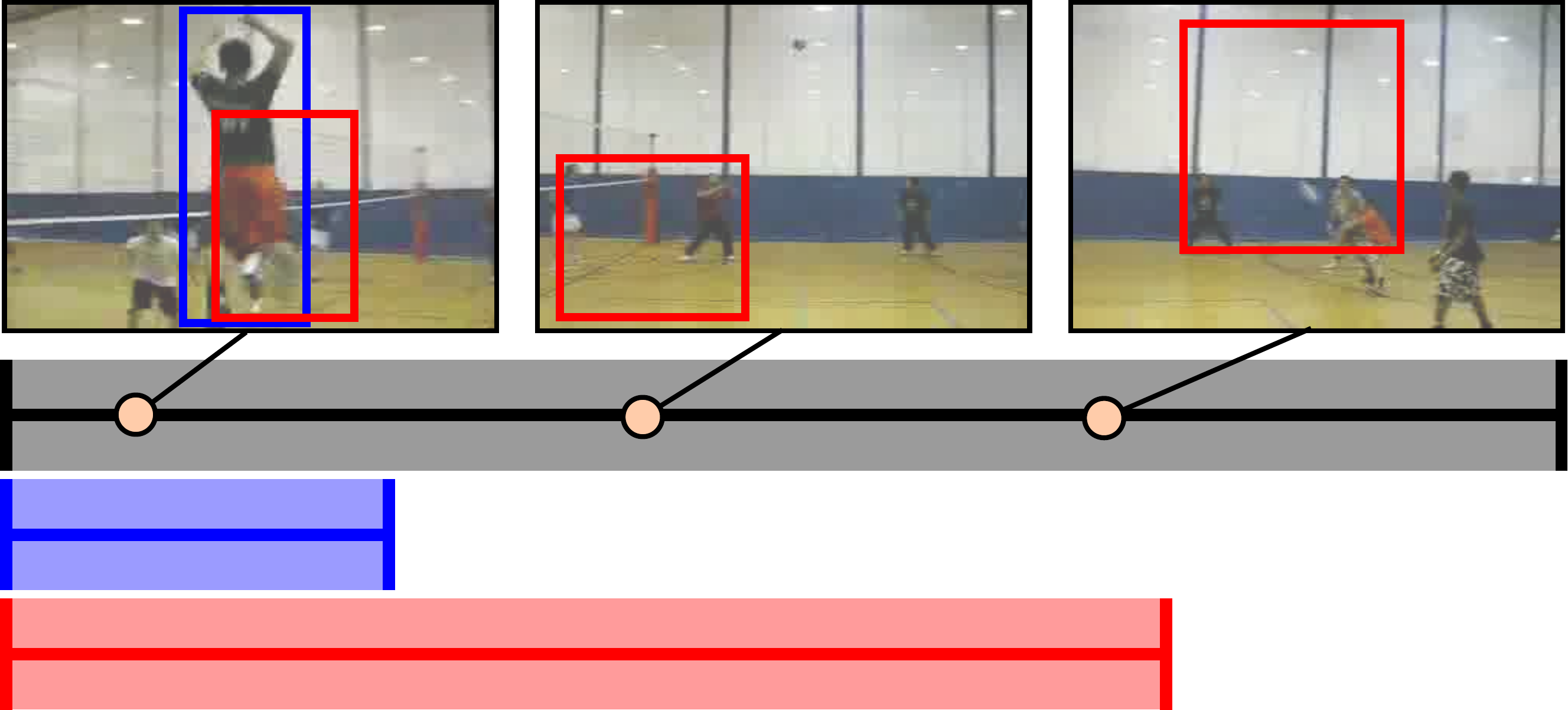}
\caption{\emph{Volleyball spike.}}
\label{fig:qual-ucf10124-c}
\end{subfigure}
\caption{\textbf{Capturing the spatio-temporal extent of actions} with STIL during training. We show three videos for UCF-101-24 with untrimmed actions. For the \emph{Golf swing} and \emph{Salsa spin}, our approach (red) correctly learns when and where the ground truth action (blue) occurs in the videos.
For \emph{Volleyball spike}, our approach fails to capture the precise nuance of spiking and learns a more general notion of playing volleyball instead.}
\label{fig:qual-ucf10124}
\end{figure*}

In Table~\ref{tab:exp1-1}, we provide an overview of the mean Average Precision scores at an overlap of 0.5. On all datasets, we observe that standard MIL can not directly compete, as it is restricted to using a single box per video as positive. The Generalized MIL baselines outperform MIL, which is a direct result of the higher number of positive box examples used per video. However, the best Generalized MIL results are outperformed by STIL. Generalized MIL yields an MAP score of 0.28, 0.10, and 0.02 on UCF Sports, J-HMDB, and UCF-101 respectively at an overlap threshold of 0.5 Our approach improves the results to 0.72, 0.30, and 0.08. These results indicate the importance of spatio-temporal constraints and initialization from zero-shot priors for action localization from video labels. In Figure~\ref{fig:qual}, we provide qualitative examples of our localization results.
\\\\
\noindent
\textbf{The impact of Temporal Prim.}
We investigate the importance of Temporal Prim in STIL. We compare to a baseline version of our approach without spatio-temporal linking.
The sole difference between our approach and the baseline is the determination of which boxes to use; either with spatio-temporal constraints modeled through Temporal Prim or without spatio-temporal constraints (\ie using all box proposals in each training video).

The comparison is shown in Table~\ref{tab:exp1-1}. Across all three datasets, we observe the benefit from linking by Temporal Prim. This result shows the importance of incorporating spatio-temporal constraints in instance learning for action localization. Moreover, incorporating the spatio-temporal constraints results in a significantly faster evaluation, because only the boxes in the tubes need to be evaluated, rather than all boxes in the video. On UCF Sports, for example, this reduces the average number of boxes to evaluate to 325, compared to 19,500 without the spatio-temporal constraints.
To highlight that our approach is not limited to trimmed videos or fixed temporal intervals, we show three qualitative examples of detected actions in untrimmed actions from UCF-101-24 in Figure~\ref{fig:qual-ucf10124}.
\\\\
\noindent
\textbf{STIL on boxes versus MIL on tubes.}
So far, we have shown the benefit of STIL over (Generalized) MIL when using box proposals for action localization. In the third ablation study, we evaluate the performance of STIL on box proposals over the current standard in weakly-supervised action localization, namely MIL on spatio-temporal tube proposals~\cite{mettes2017localizing,sapienza2012learning}. We perform this experiment on both UCF Sports and UCF-101, as comparisons are provided for these datasets. We compare our approach to the MIL baseline both with and without prior information, using the method of Mettes \etal~\cite{mettes2017localizing}.
The results are shown in Figure~\ref{fig:exp1-3} across all overlap thresholds, using mAP. For UCF Sports, we observe improvements across all overlap thresholds, especially for the high ones. At an overlap of 0.5, we obtain an mAP of 0.72, where the best baseline obtains an mAP of 0.19. On UCF-101-24, we obtain an mAP@0.5 of 0.08, compare to 0.03 of the best baseline. We conclude that STIL with spatial box proposals is preferred over MIL with spatio-temporal tube proposals.
\begin{figure}[t]
\centering
\includegraphics[width=0.825\linewidth]{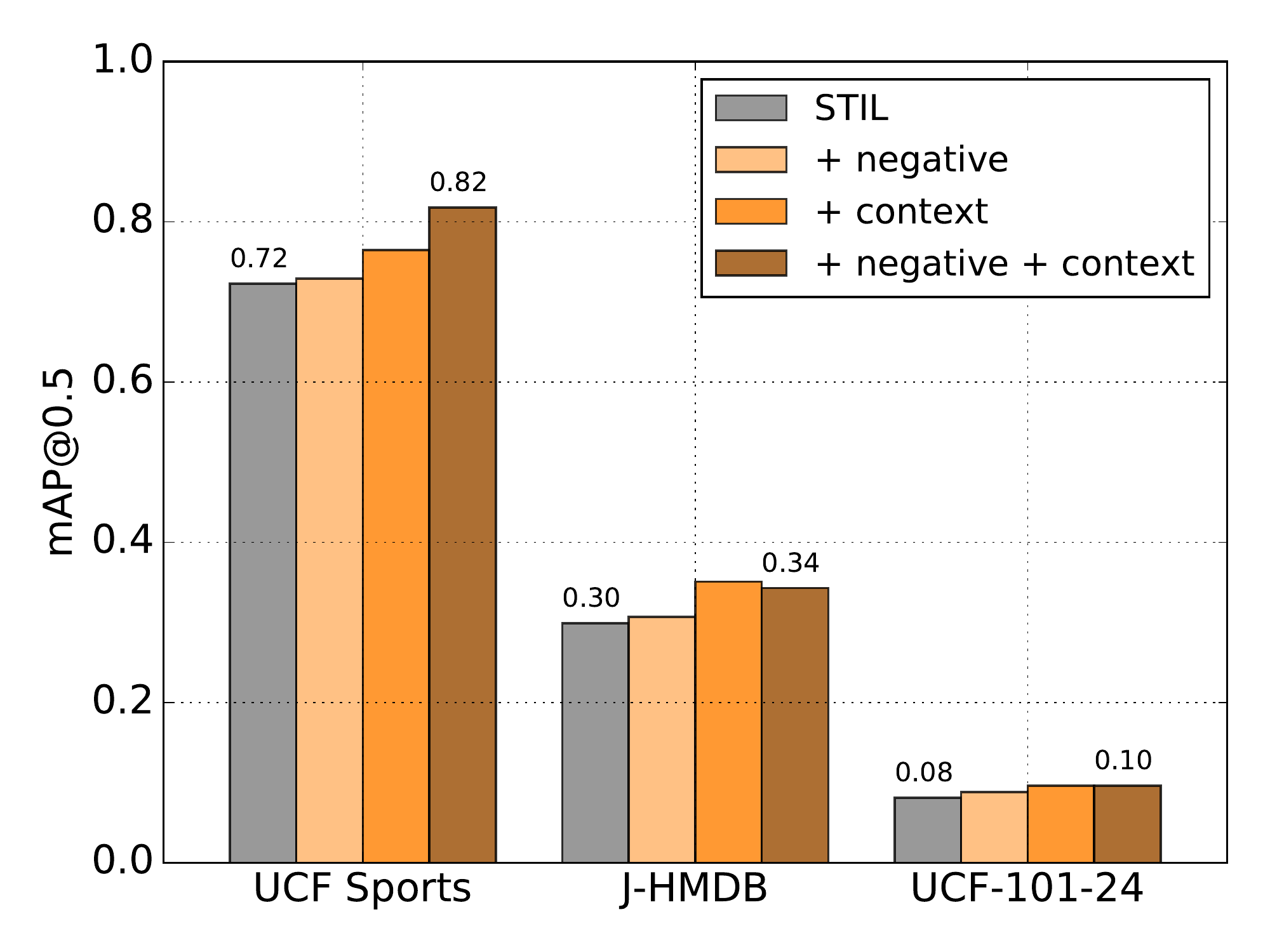}
\caption{\textbf{The impact of reranking} on the action localization performance for UCF Sports, J-HMDB, and UCF-101-24. Both negative and contextual reranking are beneficial for action localization.}
\label{fig:exp1-4}
\end{figure}
\\\\
\noindent
\textbf{The impact of reranking.}
Finally, we evaluate the effect of the global and negative reranking functions. We perform this experiment on UCF Sports, J-HMDB, and UCF-101-24. In Figure~\ref{fig:exp1-4}, we show the mAP@0.5 for our approach with and without reranking. On UCF Sports, we observe that both the negative and contextual reranking positively impact the performance, while their combination provides a further boost to 0.82, from the 0.72 without reranking. On J-HMDB, we also observe improvements for the individual rerankings, but their combination does not directly provide a further boost. Lastly on UCF-101-24, our results improve from 0.08 to 0.10. We conclude that reranking aids action localization and we will use it for our comparison to other works.

\begin{table}[t]
\centering
\resizebox{\columnwidth}{!}{%
\begin{tabular}{lrrrr}
\toprule
 & \multicolumn{2}{c}{\textbf{\footnotesize{UCF Sports}}} & \textbf{\footnotesize{UCF-101-24}} & \textbf{\footnotesize{H2T}}\\
 & AUC & mAP & mAP & mAP\\
\midrule
Sharma \etal~\cite{sharma2015action} per~\cite{li2018videolstm} & - & - & $^{\star}$0.01 & -\\
Cinbis \etal \cite{cinbis2017weakly} per \cite{mettes2017localizing}  & $^{\star}$0.14 & 0.04 & 0.04 & $^{\star}$0.00\\
Chen and Corso \cite{chen2015action} & 0.34 & - & - & -\\
Li \etal \cite{li2018videolstm} & - & - & $^{\star}$0.06 & -\\
Mettes \etal \cite{mettes2017localizing} & $^{\star}$0.32 & 0.38 & 0.06 & $^{\star}$\textbf{0.01}\\
\rowcolor{Gray}
\emph{This paper} & \textbf{0.55} & \textbf{0.82} & \textbf{0.10} & \textbf{0.01}\\
\bottomrule
\end{tabular}%
}
\caption{\textbf{State-of-the-art comparison of action localization from video labels} on three datasets. The numbers with $\star$ are provided by the authors. On UCF Sports and UCF-101-24, we improve over current approaches, while on Hollywood2Tubes all existing weakly-supervised approaches struggle to localize actions.
}
\label{tab:sota}
\end{table}

\begin{figure*}[t]
\includegraphics[width=\textwidth]{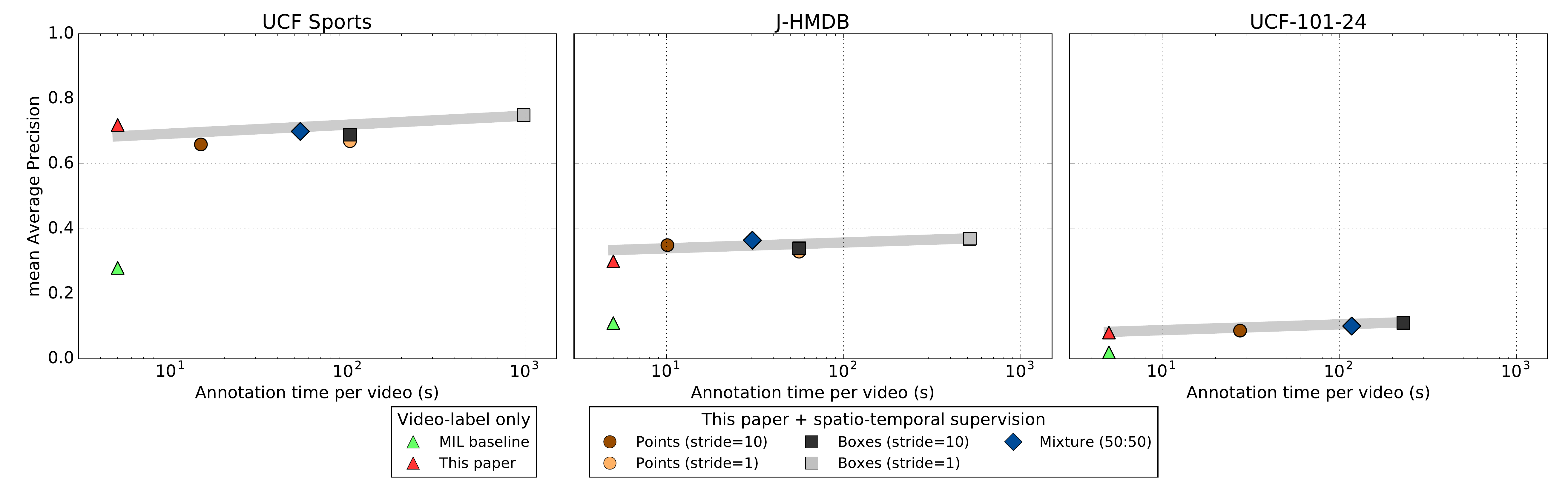}
\caption{\textbf{Benchmarking supervision levels and mixtures} for action localization with Spatio-Temporal Instance Learning on UCF Sports, J-HMDB, and UCF-101-24. The gray line indicates the linear relation between annotation cost and localization performance. On all datasets, this relation is positive, indicating that STIL is able to learn from additional supervision. We believe that this benchmark provides a step towards determining how to annotate videos most efficiently for spatio-temporal action localization.
}
\label{fig:exp3}
\end{figure*}

\subsection{State-of-the-art comparison}

\subsubsection{Comparison to weakly-supervised methods}
We have performed a comparison to the state-of-the-art in action localization using video labels only. This evaluation is performed on UCF Sports, UCF-101-24, and Hollywood2Tubes, since weakly-supervised baselines exist for these datasets.
In Table~\ref{tab:sota}, we provide an overview of all comparisons. For UCF Sports, our approach improves over the current state-of-the-art in weakly-supervised action localization both for the AUC and mAP metrics. We obtain an AUC score of 0.55 and mAP of 0.82 at an overlap of 0.5, compared to 0.32 and 0.38 for the highest performing method on this dataset~\cite{mettes2017localizing}.
For UCF-101-24, our approach outperforms Li \etal~\cite{li2018videolstm} and Mettes \etal~\cite{mettes2017localizing}. At an overlap threshold 0.5, we obtain an mAP of 0.10, compared to 0.06 for both other approaches.
We have also performed a comparison on Hollywood2Tubes. This dataset is multi-label temporally untrimmed, akin to other recent datasets such as AVA~\cite{gu2018ava}, and DALY~\cite{weinzaepfel2017human}, and VIRAT~\cite{oh2011large}, but Hollywood2Tubes provides comparisons to other weakly-supervised methods. We obtain a low mAP of 0.01, similar to other weakly-supervised approaches, indicating that action localization in fully untrimmed multi-label videos is an open action localization problem. Nonetheless, our approach does provide state-of-the-art results for action localization from video labels.

\subsubsection{Comparison to other supervision levels}
While the focus of Spatio-Temporal Instance Learning is on action localization from video labels, it can also incorporate additional supervision. This enables an apples-to-apples comparison of supervision levels and even mixtures thereof using the same underlying features and optimization. As a final experiment, we benchmark the action localization performance as a function of the annotation cost for four annotation strategies and six supervision variants: (1) using video labels only, (2) point annotations each frame, (3) point annotations each tenth frame, (4) box annotations each frame, (5) box annotations each tenth frame, and (6) a 50/50 mixture of the first and fifth variant. For all variants, we have investigated the estimated annotation time per video.
A detailed outline of the annotations costs is provided in the appendix.

In Figure~\ref{fig:exp3}, we show the results on UCF Sports, J-HMDB, and UCF-101-24. On all datasets, we observe a positive relation between the amount of supervision and the localization performance. On UCF Sports, adding supervision with sparse points or box annotations does not improve the results. Only with full box supervision, the results improve to 0.75 from 0.72. On J-HMDB and UCF-101-24, any additional supervision has a positive influence on the performance. On J-HMDB, the supervision mixture provides a high performance at a modest additional annotation cost. On UCF-101-24, the relative improvement from additional supervision is largest of all datasets (from 0.08 with video labels to 0.11 with full supervision).
We note that the comparison is meant to compare supervision levels within the same model, not to obtain optimal results for each level separately.
The comparison naturally shows that supervision helps performance. More interestingly, we believe it is a step towards the research question: what annotation strategy for spatio-temporal action localization provides maximal performance with minimal annotation effort?


\section{Conclusions}
We propose Spatio-Temporal Instance Learning, a model for spatio-temporal action localization from video class labels only.
We introduce three conditions, an objective function, and optimization to perform the instance learning. Furthermore, we propose an efficient linking algorithm and two reranking strategies to further improve the localization results.
Experimental evaluation shows the effectiveness of our proposal, resulting in state-of-the-art weakly-supervised action localization.
The experiments also pinpoint a number of open challenges in spatio-temporal action localization from video labels. First, distinguishing multiple actions in the same video (\eg \emph{Catch} and \emph{Throw}) is problematic given the lack of spatio-temporal information about the actions.
Second, large-scale action localization on datasets such as Sports1M~\cite{karpathy2014large} and Youtube8M~\cite{abu2016youtube} becomes a possibility when only video labels are required during training.
Lastly, the comparison of supervision levels opens up to the question of how, when, and where videos should be annotated for maximal performance with minimal annotation effort. This holds especially for recent untrimmed multi-label datasets, where weakly- and strongly-supervised action localization have yet to make an impact.

\appendix
\section{Annotation cost per supervision level}
Here, we outline how we derived the annotation costs for the different supervision levels for the final benchmark experiment of our main paper.
Following~\cite{bearman2016s}, we make the annotation cost of the different supervision levels modular, \ie components are measured separately and summed where appropriate.

Retrieving the action class label per video on both datasets takes roughly 5 seconds. We attribute this time due to the low cardinality of the set of actions. Both datasets have few action classes (10 for UCF Sports, 21 for J-HMDB, 24 for UCF-101-24) and only a single action occurs per video, making a selection relatively easy.

Annotating a bounding box in a video frame takes an estimated 15 seconds. This estimate is lower than the estimated box annotation time on ImageNet~\cite{su2012crowdsourcing} (34.5 seconds per box per image), which is because we have sequences of images, rather than individual images. Hence, we have a prior location of actions from previous frames to speed up the box annotation. The estimation is closer to the one made by Russakovsky \etal~\cite{russakovsky2015best} (10 to 12 seconds).

Annotating a point takes an estimated 1.5 seconds in a single frame. This estimate is in line with Mettes \etal~\cite{mettes2016spot}, who estimated that points are ten times faster to annotate than boxes. The estimate also  aligns with point supervision in image segmentation, which takes 2.4 seconds for the first instance in an image and 0.9 seconds for each consecutive instance in the same image~\cite{bearman2016s}. Our estimate falls within this range.

Based on these estimates, we provide the overall annotation costs per video in seconds in Table~\ref{tab:costs} for UCF Sports, J-HMDB, and UCF-101-24. These are the estimates used in our final benchmark experiment.

\begin{table}[h]
\centering
\resizebox{\columnwidth}{!}{%
\begin{tabular}{lrrr}
\toprule
\textbf{Supervision level} & \multicolumn{3}{c}{\textbf{Annotation cost (sec.)}}\\
& UCF Sports & J-HMDB & UCF-101-24\\
\midrule
Video labels & 5.00 & 5.00 & 5.00\\
Points (stride=10) & 14.75 & 10.10 & 27.50\\
Mixture (50:50) & 53.75 & 30.50 & 117.50\\
Points (stride=1) & 102.50 & 56.00 & -\\
Boxes (stride=10) & 102.50 & 56.00 & 230.00\\
Boxes (stride=1) & 980.00 & 515.00 & -\\
\bottomrule
\end{tabular}%
}
\caption{\textbf{Annotation cost per supervision level} measured in seconds per video on both UCF Sports, J-HMDB, and UCF-101-24. The overall costs are higher for UCF Sports since the videos are longer on average than J-HMDB (65 frames per video vs. 34 frames per video for J-HMDB). For UCF-101-24, a stride of 1 is not employed, as features are extracted every fifth frame.}
\label{tab:costs}
\end{table}

\appendix

\vspace{8mm}
{
\small
\noindent
\textbf{Acknowledgments.}
Supported by the Intelligence Advanced Research Projects Activity (IARPA) via Department of Interior/Interior Business Center (DOI/IBC) contract number D17PC00343. The U.S. Government is authorized to reproduce and distribute reprints for Governmental purposes notwithstanding any copyright annotation thereon. Disclaimer: The views and conclusions contained herein are those of the authors and should not be interpreted as necessarily representing endorsements, either expressed or implied, of IARPA, DOI/IBC, or the U.S. Government.
}

{\small
\bibliographystyle{ieee}
\bibliography{stil-cvpr19}
}

\end{document}